\documentclass{article}

\usepackage[usenames,dvipsnames]{color}
\usepackage{multirow}
\usepackage{verbatim}

\newcommand{\BE}{\mathbb{E}}
\newcommand{\mc}{\mathcal}
\usepackage{amsthm}
\usepackage{graphicx}

\makeatletter
\newcommand{\printfnsymbol}[1]{
  \textsuperscript{\@fnsymbol{#1}}
}
\makeatother
\PassOptionsToPackage{sort&compress,numbers}{natbib}
     \usepackage[preprint]{neurips_2020}

\usepackage[utf8]{inputenc}
\usepackage[table,xcdraw]{xcolor}

\usepackage{physics}
\usepackage[T1]{fontenc}
\usepackage{hyperref}
\newcommand{\pdfcolor}{blue}
\hypersetup{
    colorlinks=true,
    linkcolor=\pdfcolor,
    citecolor=\pdfcolor,
    urlcolor=blue}

\usepackage{url}
\usepackage{booktabs}
\usepackage{amsfonts}
\usepackage{nicefrac}
\usepackage{microtype}
\usepackage{amsmath}
\usepackage{algorithm,algorithmic}
\usepackage{graphicx}

\title{Generalization by Recognizing Confusion}

\author{
  Daniel Chiu\thanks{These authors contributed equally to the work.} \\
  Department of Computer Science\\
  Harvard University\\
  Cambridge, MA 02183 \\
  \texttt{chiud@college.harvard.edu} \\
\And
  Franklyn H.~Wang\printfnsymbol{1} \\
  Department of Mathematics\\
  Harvard University\\
  Cambridge, MA 02183 \\
  \texttt{franklyn\_wang@college.harvard.edu} \\
\And
  Scott Duke~Kominers \\
  Harvard Business School\\ 
  Boston, MA 02163 \\
  \texttt{kominers@fas.harvard.edu} \\
}

\begin{document}

\maketitle

\begin{abstract}
A recently-proposed technique called self-adaptive training augments modern neural networks by allowing them to adjust training labels on the fly, to avoid overfitting to samples that may be mislabeled or otherwise non-representative. By combining the self-adaptive objective with mixup, we further improve the accuracy of self-adaptive models for image recognition; the resulting classifier obtains state-of-the-art accuracies on datasets corrupted with label noise. Robustness to label noise implies a lower generalization gap; thus, our approach also leads to improved generalizability. We find evidence that the Rademacher complexity of these algorithms is low, suggesting a new path towards provable generalization for this type of deep learning model. Last, we highlight a novel connection between difficulties accounting for rare classes and robustness under noise, as rare classes are in a sense indistinguishable from label noise. Our code can be found at~\href{https://github.com/Tuxianeer/generalizationconfusion}{\color{magenta} \texttt{https://github.com/Tuxianeer/generalizationconfusion}}.
\end{abstract}

\section{Introduction}

Most modern machine learning algorithms are trained to maximize performance on meticulously cleaned datasets; relatively less attention is given to robustness to noise. Yet many applications have noisy data, and even  curated datasets like ImageNet have possible errors in the training set \cite{russakovsky2015imagenet}.

Recently, Huang et al.~\cite{huang2020self} introduced a framework called \textit{self-adaptive training} that achieves unprecedented results on noisy training data. Self-adaptive training augments an external neural network through combining two procedures: it adjusts training labels that the model concludes are likely to be inaccurate, and it gives lower weight to training examples for which it is most uncertain. Intuitively, self-adaptive models are in some sense able to recognize when they are ``confused'' about examples in the training set. They fix the labels of confusing examples when they are convinced that the confusion is likely to be the result of incorrect labels; meanwhile, they down-weight the examples that do not seem to match any of the classes well.

We develop some theory for self-adaptive training and find that the re-weighting process should perform best when the weights accurately reflect in-distribution probabilities, which occurs precisely when the underlying model is well-calibrated. This leads us to augment \cite{huang2020self} with a calibration process called mixup, which feeds models examples from a derived continuous distribution. 

Our \textit{self-adaptive mixup} framework combines the label noise robustness of self-adaptive training with the smoothing nature of mixup. In experiments on CIFAR10 and CIFAR100, self-adaptive mixup outperforms the previous noisy-label state-of-the-art under almost all levels of noise.

Additionally, we find that self-adaptive mixup generalizes especially well, with generalization gaps an order of magnitude below those of standard neural networks. These findings are consistent with the theoretical result from \cite{shalev2014understanding} that high accuracy under label noise implies a lower generalization gap. And we might intuitively expect that self-adaptive methods should generalize well, as they in some sense focus on the most ``representative'' datapoints---the data points about which they have high confidence early on---and place less weight on outliers that other models overfit. 

Building on the empirical generalization performance of our self-adaptive framework, we examine several properties of the method that we hope may provide a path to a formal proof of generalization. We show experimentally that unlike standard neural networks, self-adaptive frameworks do not fit random labels---in fact, after label correction, the model's posterior class-conditional probabilities are all essentially $1/\#\text{classes}$, irrespective of the initial labels. This shows that the experimental Rademacher complexity of self-adaptive training may be low; a corresponding theoretical result, if found, would directly imply a bound on the generalization gap. Additionally, self-adaptive methods do not fully fit the training data, which implies that the generalization gap (between training error and test error) can be closed from both sides. Finally, self-adaptive training does not fit rare classes when the distribution is imbalanced. Although this at first seems like a deficiency, we note that it is a necessary consequence of robustness to label noise---and also implies a certain form of robustness to overfitting to outliers, which is key to generalization.

Taken together, our results on generalization point to the possibility that self-adaptive models' ability to work around confusing examples in fact drives strong generalization performance. Indeed, their low Rademacher complexity arises precisely because they recognize completely noisy data sets as especially confusing and avoid conjuring faulty patterns. The other properties we highlight arise because some examples in the training data are more confusing than others, and self-adaptive models treat those examples as unlikely to be representative of their labeled classes.
 
Overall, our work highlights the power of the self-adaptive framework and presents direct improvements on the state-of-the-art for classification and generalization under label noise. We also give conceptual and empirical support for the possibility that self-adaptive frameworks may eventually admit a full proof of generalization---a type of argument that has remained elusive for neural networks.

\subsection{Related Work}

\paragraph{Learning from Noisy Labels.} At a high level, there are two prior classes of approaches for learning from noisy labels: \emph{label correction} attempts to fix the labels directly \cite{tanaka2018joint,thulasidasan2019combating,nguyen2020self}; \emph{loss correction} attempts to fix the loss function through methods like backward and forward correction (see, e.g., \cite{DBLP:conf/cvpr/PatriniRMNQ17}). Label correction, however, tends to be slow in practice because they involve multiple rounds of training. And as \cite{lukasik2020does} recently highlighted, loss correction may have significant calibration errors. 

Self-adaptive training \cite{huang2020self} elegantly combines both approaches, altering the loss function and fixing the labels endogenously during training. The approach we introduce here strengthens the loss function correction of the self-adaptive approach by bringing the weights used in the model closer to the theoretical ideal. (See Tables~\ref{table:gengaps} and~\ref{table:gengaps100} for experimental comparisons.)

\paragraph{Theoretical Advances in Generalization.}
Showing uniform convergence is a classic approach to proving generalization, but \cite{DBLP:conf/nips/NagarajanK19} demonstrated that it cannot be proven for standard neural networks. Nevertheless, \cite{DBLP:conf/nips/NagarajanK19} left the door open to the possibility that uniform convergence could be shown for neural networks that are trained by different approaches, such as those described in this paper.

While we are not able to give a full proof of generalization for self-adaptive models in this work, we gesture in that direction by providing both theoretical and empirical evidence that our models generalize in a way that might be provable. In particular, we give evidence that proving generalization of self-adaptive models may be easier than for standard neural networks (see Section~\ref{sec:confgen}).

\paragraph{Influence Functions.} Last, our work is to some degree related to the pathbreaking methods of \cite{DBLP:conf/icml/KohL17}, who used ``influence functions'' to estimate the impact of each data point on model predictions. This method reveals the data points with the largest impact on the model, and in the case of noisy labels, \cite{DBLP:conf/icml/KohL17} showed that by manually correcting the labels of these most-relevant data points one can substantially improve performance. The self-adaptive approach in some sense employs a similar feedback process, but without the need for a human in the loop. 

\section{Self-Adaptive Training}

\paragraph{Motivation.}
Incorrect data, in the form of label noise, is a significant problem for machine learning models in practice. Neural networks are robust to small numbers of faulty labels, but drop in performance dramatically with, say, 20\% label noise. Self-adaptive training, proposed by \cite{huang2020self}, is an augmentation that can be performed atop any training process to increase robustness to label noise. After a fixed number of start-up epochs, the model begins adjusting the labels of each example in the training set towards the model's own predictions. In addition, rather than being given equal weight, each example in the training set is \emph{confidence-weighted}: that is, each example's contribution to the loss function is weighted by the maximum predicted class probability---a well-known metric for a model's confidence about an example's classification \cite{hendryks2016baseline}. Intuitively, a self-adaptive model recognizes when it is ``confused'' by an example and proceeds to down-weight that example's contribution to the loss function, with the eventual goal of correcting the labels that seem most likely to be noisy.

\paragraph{The Algorithm.}

Self-adaptive training modifies regular training in two ways:

\begin{itemize}
    \item \textsl{Label Correction:} During each iteration of training, the model updates its stored labels of the training data (known as ``soft labels'' and denoted $\mathbf{t}_i$ for example $i$) to be more aligned with the model's predictions. 
    \item \textsl{Re-weighting:} The $i$'th sample is weighted by $w_i$, the maximum predicted class probability, to reduce the weight of examples less likely to be in-distribution.
\end{itemize}

To gradually move the soft labels towards the predictions, the model uses a \emph{momentum update}: $\mathbf{t}_i \leftarrow \alpha \times \mathbf{t}_i + (1 - \alpha) \times \mathbf{p}_i$. Intuitively, when $\alpha = 1$ the soft label is not updated at all, and when $\alpha = 0$ the soft label is changed completely to match the model's prediction. Thus $\alpha = 1$ corresponds to regular training, and $\alpha = 0$ essentially corresponds to early stopping, because if the labels are immediately set to match the predictions, then the model will cease updating (having already minimized the loss function). Thus, letting $\alpha$ vary between $0$ and $1$ interpolates between early stopping and regular training in some sense---and may allow combining the benefits of both. 

Additionally, in each step of the algorithm, the samples' contributions to the model are re-weighted according to $w_i = \max_{j} \mathbf{t}_{i, j}$,  corresponding to a measure of confidence in its ability to classify sample $i$. If one of a sample's predicted class probabilities $\mathbf{t}_{i, j}$ is substantially higher than others ($\mathbf{t}_{i, j'}$, $j'\neq j$), then the sample is more likely to be representative of class $j$. Furthermore, especially in the context of noisy labels, a low value of $w_i$ corresponds to the model being ``confused'' about sample $i$, in the sense that it thinks that sample could come from many different classes. One source of confusion could be an out-of-distribution sample due to a noisy label, and it makes sense to assign lower weight to such possibly out-of-distribution examples.

Note that the two features of self-adaptive training are highly complementary: unlike approaches that simply discard examples with low maximum predicted class probability, self-adaptive training can eventually bring the weight for those examples back up---allowing the model to make full use of those samples if and when it becomes confident that it classifies them correctly.\footnote{This may happen after those examples themselves have been relabeled, or when enough \textit{other} examples have been relabeled.} For completeness, we reproduce the self-adaptive training algorithm of  \cite{huang2020self} as Algorithm~\ref{alg:seq} in Appendix~\ref{ap:B}. 

\subsection{Why Does Self-Adaptive Training Work?}

In self-adaptive training, the label correction step allows the model to avoid discarding samples it is confident are wrongly labeled (or worse, overfitting to them); instead, the model adjusts the labels of those samples and learns from the corrected data. The re-weighting component makes use of the fact that the model believes some samples are more likely than others to be in-distribution. Note that while self-adaptive training must determine which labels are incorrect, it takes advantage of the fact that the model's performance on individual training examples actually directly implies a set of in-distribution probabilities.

\begin{figure}[t]
\vskip -0.15in
    \centering
        \caption{Neural networks overfitting to noise (reproduced from~\cite{huang2020self}): The performance on the clean training set (the green curve) rises above the total number of correct examples in the training set (the dotted line), before the model fits the entire noisy training set (the red curve) and drops in accuracy on the clean validation set (the yellow curve).\label{fig:my_label}}
    \includegraphics[scale = 0.8]{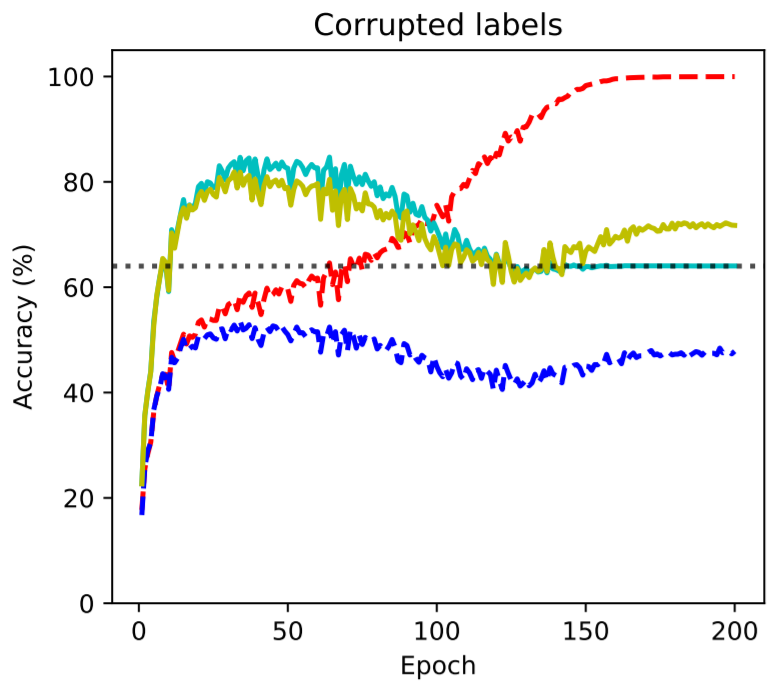}
    \vskip -0.15in
\end{figure}

A main observation of \cite{huang2020self} is that in scenarios with high amounts of label noise, modern neural networks will initially learn, or ``magnify,'' the simple patterns in the data, achieving higher accuracy on the test set than is even present in the (noisy) training set. However, in later epochs the models train to essentially 100\% accuracy on the training set, thereby overfitting to the noise (see Figure~\ref{fig:my_label}, which we reproduce from \cite{huang2020self}). Self-adaptive training mitigates this problem by correcting that noise when earlier epochs are convinced the labels are wrong.

However, the preceding explanation cannot be the entire story, as \cite{huang2020self} also found that self-adaptive training improves test accuracy on ImageNet with \textit{no label noise}. While it is possible that the base ImageNet dataset itself may have some latent label noise in the form of incorrectly classified images, we propose a more nuanced theory that expands on the idea of magnification mentioned above.

Last year, \cite{li2019gradient} showed that early stopping is provably robust to label noise---and more broadly, it is believed that the early epochs of training produce simpler models that are more easily representable by a neural network. It is natural to think that the data points that feed into those early networks are somehow more easily interpreted than others. In that case, by early-epoch label correction, self-adaptive training has the ability to down-weight confusing and difficult-to-learn examples; and by doing so, it may have greater ability to extract and learn the ``main ideas'' from a dataset.

\subsection{Optimal Re-Weighting}\label{subsec:proof}

Despite all the advantages just described, it seems that the self-adaptive training framework still leaves some room for improvement. In an ablation study, \cite{huang2020self} attributed almost all of the gains from self-adaptive training to the label correction component, while re-weighting provided only marginal further benefits. Prima facie, this is surprising because re-weighting can mask non-representative examples---especially relevant in the presence of noise. 

We now provide a novel mathematical derivation which suggests that the choice of weights used in the prior self-adaptive training setup might be preventing re-weighting from achieving its full potential. 
Formally, the goal of the model $\theta$ is to minimize the true loss \[ \mathcal{L}_{\mathcal{D}}(\theta) = \mathbb{E}_{(\mathbf{x}, y) \sim \mathcal{D}}[\ell(f_{\theta}(\mathbf{x}), y)]. \]

Now, suppose that there are many samples $\{(\mathbf{x}_i, y_i)\}_{1 \le i \le n}$, where some come from a distribution $\mathcal{D}$ and some come from a distribution $\mathcal{D}'$. Let the datapoint $(\mathbf{x}_i, y_i)$ be drawn from the distribution $\mathcal{D}$ with probability $p_i$ and $\mathcal{D}'$ with probability $1 - p_i$. 

For many kinds of noise (including uniform label noise), $\BE_{(\mathbf{x}, y) \sim \mc D'}[\ell(f_{\theta}(\mathbf{x}), y)]$ is constant for all $\theta$. Since to compare models it suffices to compute loss up to translation, we  assume that $\BE_{(\mathbf{x}, y) \sim \mc D'}[\ell(f_{\theta}(\mathbf{x}), y)]$ is always $0$.

For any weights $q_i$ ($1\le i\le n$), we can write an unbiased Monte Carlo estimator of the true loss $\mathbb{E}_{(\mathbf{x}, y) \sim \mathcal{D}}[\ell(f_{\theta}(\mathbf{x}), y)]$ as (modulo some algebra)
\begin{equation}
\frac{\sum_{i = 1}^{n} q_i \ell(f_{\theta}(\mathbf{x}_i), y_i)}{p_1q_1 + p_2q_2 + \ldots + p_nq_n}. \label{eq:est}\end{equation} 

We seek to choose the $q_i$ to minimize the variance of the estimator \eqref{eq:est}. The scaling of the $q_i$ is irrelevant, so treat $\norm{q}_2$ as constant. Under the assumption that the variance of $\ell(f_{\theta}(\mathbf{x}_i), y_i)$ is $v$ for all $i$, and since the pairs $(\mathbf{x}_i, y_i)$ are independent, the variance of the numerator of \eqref{eq:est} is just $v\cdot \norm{q}_2^2$---a constant. Thus, minimal variance is achieved in \eqref{eq:est} when the denominator is maximized. By the Cauchy-Schwarz inequality, for fixed $\norm{q}_2$, the denominator is maximized with $q_i \propto p_i$. 

The preceding derivation shows that in theory the optimal weights are proportional to the in-distribution probabilities of each point. The formulation of self-adaptive training by \cite{huang2020self} uses as weights the maximum predicted class probability. \cite{hendryks2016baseline} suggests that the magnitudes of these weights generally increase with in-distribution probabilities, but that the relationship is not linear. As a consequence, we might be able to further improve the self-adaptive model by tuning the weights to be more proportional to the in-distribution probabilities; we do this in the next section.

\section{Calibrating Self-Adaptive Training}

To get closer to optimal weights in self-adaptive training, we draw upon a class of methods that have been developed to better quantify predictive uncertainty. Specifically, a model is said to be \textit{calibrated} if when it suggests a sample has probability $p$ of being of class $c$, the correct posterior probability of class $c$ is indeed $p$---exactly the property we show in Subsection~\ref{subsec:proof} to be necessary for optimal re-weighting in self-adaptive training.

As noted by \cite{guo2017calibration}, modern neural networks tend to be poorly calibrated out-of-the-box. And---recalling our analysis from the previous section---if the model underlying our self-adaptive training process is poorly calibrated, then the weights it produces will not match the theoretical ideal. We thus pair self-adaptive training with a calibration method called \textit{mixup} \cite{DBLP:conf/iclr/ZhangCDL18}. As we show in experiments, using mixup to improve the weights of self-adaptive training yields substantial improvements.

\paragraph{Self-Adaptive Mixup.}

\textit{Mixup} is a data augmentation procedure proposed by \cite{DBLP:conf/iclr/ZhangCDL18} that is known to improve calibration (\cite{DBLP:conf/nips/ThulasidasanCBB19}) when applied to standard models. Under mixup, instead of training on the given dataset $\mathcal{D}$, the model trains on a ``mixed up'' dataset, draws from which correspond to linear combinations of pairs of datapoints in $\mathcal{D}$. 

Formally, given a dataset of images and labels $\mathcal{D}=\{\mathbf{x}_i, \mathbf{t}_i\}$ and a smoothing parameter $\alpha$, mixup considers convex combinations 
\begin{align*}
    \widetilde{\mathbf{x}} &:= \lambda \mathbf{x}_i + (1 - \lambda) \mathbf{x}_j \\ 
    \widetilde{\mathbf{t}} &:= \lambda \mathbf{t}_i + (1 - \lambda) \mathbf{t}_j,
\end{align*}
where $\lambda \sim \text{Beta}(\alpha, \alpha)$. Loss is normally computed via the standard cross-entropy loss on $\widetilde{\mathbf{x}}$ and $\widetilde{\mathbf{t}}$.

In the paper introducing mixup (\cite{DBLP:conf/iclr/ZhangCDL18}), the authors interpreted the approach as encouraging the model to linearly interpolate between training examples, as opposed to training on only extreme ``all or nothing'' examples. In the context of calibration, this alleviates model overconfidence by teaching the model to output non-binary predictions.

While mixup applies out-of-the-box to a standard training process, the two main aspects of self-adaptive training---reweighting and label correction---both potentially conflict with mixup. We thus develop the \textit{self-adaptive mixup} algorithm, which integrates mixup into the self-adaptive paradigm. Self-adaptive mixup takes in two parameters: the $\alpha$ used by standard mixup, and a ``cutoff'' parameter $\gamma$. As in self-adaptive training, the algorithm maintains a set of soft labels for the \textit{original} training examples. The two components of self-adaptive training are then adjusted as follows:
\begin{itemize}
 \item \textsl{Label Correction:} Self-adaptive training involves updating soft labels of training examples toward model predictions---but mixed-up models do not train on examples in the original training set directly. Thus, we need a rule that determines when to update the soft labels; intuitively, we choose to update only when training on a mixed-up example that is sufficiently similar to an original example. Specifically, the algorithm only updates soft labels when training on mixed-up examples that are at least a $(1 - \gamma)$ proportion of a single original example: i.e., for a mixed-up example $\mathbf{x'} = \lambda\mathbf{x}_i + (1-\lambda)\mathbf{x}_j,$ we update the soft label of $\mathbf{x}_i$ iff $\lambda > 1 - \gamma$, and update the soft label of $\mathbf{x}_j$ iff $\lambda < \gamma$. 
 \item \textsl{Re-weighting:} The reweighting component of self-adaptive training weights the contribution of each example to the loss by the maximum class-conditional probability predicted by the model. Analogously, in self-adaptive mixup, we weight each mixed-up example's contribution to the loss by the maximum class-conditional probability the model assigns to that mixed-up example.
\end{itemize}

\paragraph{Experiments.}

We present the results of experiments testing self-adaptive mixup. We performed experiments on the CIFAR10 and CIFAR100 datasets with uniform label noise rates of $r \in \{0.2,0.4,0.6,0.8\}$. All experiments were conducted with a Wide Resnet-34x10 \cite{zagoruyko2016wide}, following~\cite{huang2020self}. Each individual run took approximately 4 hours to run on one TPU v2. 

The results are displayed in the bottom two rows of Table~\ref{tab:noisycifarresults}. Again following \cite{huang2020self}, we include for comparison a number of past results of models trained for label noise, including self-adaptive training and (standard) mixup. In self-adaptive mixup, we fix $\gamma = 0.1$ and use two values of $\alpha$, $0.2$ and $1.0$. For these hyperparameters, $0.1$ was chosen intuitively as a low but reasonable value, and the two values of $\alpha$ were suggested by past work on mixup \cite{DBLP:conf/iclr/ZhangCDL18}.\footnote{Due to lack of computing power, we were only able to test self-adaptive mixup with $\gamma = 0.1$; we suggest exploring optimizing this hyperparameter as a potential direction for future research.} Finally, each reported self-adaptive mixup accuracy is the median result across three independent runs with different noisy labels and random seeds.

\begin{table*}[t!]
\vskip -0.15in
\caption{Test Accuracy (\%) on CIFAR10 and CIFAR100 datasets with various levels of uniform label noise injected into the training set. We compare self-adaptive mixup (``Ours'') with previous work under identical experiment settings. The best predictive accuracies are highlighted in \textbf{bold}.}
\label{tab:noisycifarresults}
\begin{center}
\begin{small}
\begin{sc}
\vskip -0.05in
\begin{tabular}{lcccccccc}
\toprule
 & \multicolumn{4}{c}{CIFAR10} & \multicolumn{4}{c}{CIFAR100} \\
\midrule
\multirow{2}{*}{Method} & \multicolumn{4}{c}{Label Noise Rate} & \multicolumn{4}{c}{Label Noise Rate} \\
        & 0.2 & 0.4 & 0.6 & 0.8 & 0.2 & 0.4 & 0.6 & 0.8   \\
\midrule
CE + Early Stopping~\cite{huang2020self}                        & 85.57 & 81.82 & 76.43 & 60.99 & 63.70 & 48.60 & 37.86 & 17.28 \\
Mixup~\cite{DBLP:conf/iclr/ZhangCDL18}                 & 93.58 & 89.46 & 78.32 & 66.32 & 69.31 & 58.12 & 41.10 & 18.77 \\
SCE~\cite{wang2019symmetric}                 & 90.15 & 86.74 & 80.80 & 46.28 & 71.26 & 66.41 & 57.43 & 26.41 \\
SAT~\cite{huang2020self}                    & 94.14 & 92.64 & 89.23 & 78.58 & 75.77 & 71.38 & 62.69 & 38.72 \\
SAT + SCE~\cite{huang2020self}                    & 94.39 & 93.29 & 89.83 & 79.13 & 76.57 & 72.16 & 64.12 & \textbf{39.61} \\
 \midrule
Ours ($\alpha = 0.2$)                                    & 94.83 & 93.72 & \textbf{91.21} & \textbf{80.25} & 75.21 & 72.45 & \textbf{65.12} & 38.96 \\
Ours ($\alpha = 1.0$)                                    & \textbf{95.48} & \textbf{94.15} & 89.31 & 74.45 & \textbf{78.03} & \textbf{72.67} & 62.59 & 32.65 \\
\bottomrule
\end{tabular}
\end{sc}
\end{small}
\end{center}
\vskip -0.25in
\end{table*}

\paragraph{Discussion.}

As we see in Table~\ref{tab:noisycifarresults}, self-adaptive mixup improves on the state-of-the-art in all but one combination of dataset and noise rate---and the improvement is often substantial. 

We see also that as the noise rate increases, the optimal choice of $\alpha$ decreases. To see why this might be, note that the mixup ratio is drawn from $\text{Beta}(\alpha, \alpha)$, which approaches $\text{Bern}(0.5)$ as $\alpha \to 0$ and approaches $1/2$ as $\alpha \to \infty$. Recall that label correction happens only when the drawn ratio is not between $0.1$ and $0.9$. If $p_\alpha$ is the probability density function of $\text{Beta}(\alpha, \alpha)$, labels are updated with probability $1 - \int_{0.1}^{0.9} p_\alpha(x)\, dx$, which decreases with $\alpha$. Thus, we conjecture that the observed phenomenon occurs because labels are updated via label correction more often when $\alpha$ is low, and frequent label correction is more important when labels are more noisy ex ante. 

The dependence of the optimal $\alpha$ on the noise rate is potentially problematic in practice, as the noise rate in real-world datasets is of course not known a priori. However, we believe that noise rates in the wild should be substantially below one half, and thus would recommend $\alpha = 1.0$ in general.\footnote{Furthermore, note that the self-adaptive approach actually exposes information about the noise rate of the dataset via the frequency of changed labels---which suggests an adaptive approach that varies $\alpha$ and/or $\gamma$ during training may be fruitful.}

We further note that the train and test performances of self-adaptive mixup are much closer to each other than those of self-adaptive training. This suggests---as we discuss this further in the next section---that self-adaptive mixup may submit to guarantees on generalization performance.

\section{Noticing Confusion as a Path to Generalization}\label{sec:confgen}

In classification, the central metric of interest is the \emph{generalization error}, which measures how the model performs on the true distribution, as proxied by error on a test set \cite{shalev2014understanding}. A model's \textit{generalization gap} is the absolute difference between its training error and its generalization error. The self-adaptive training models of \cite{huang2020self} perform especially well on this metric; \cite{huang2020self} found generalization gaps for their methods to be substantially lower than those of standard neural networks. 

Relatedly, \cite{shalev2014understanding} notes a connection between generalization and stability of models under perturbations to the training set: they consider the effect of replacing a sample from the training set, and note that the difference in the probability of a model classifying the sample correctly between the case in which (1) the model has the sample and the case in which (2) it is replaced is equal to the generalization gap of the model. We consider the setting where data points are not replaced but instead given random labels, which is strictly more difficult. In this setting, better robustness to label noise---such as that of self-adaptive mixup---is strong evidence for low generalization error. For instance, Table~\ref{table:gengaps} suggests that self-adaptive mixup models have \emph{even lower} generalization gaps than vanilla self-adaptive training.

\begin{table*}[t!]
\vskip -0.15in
\caption{Generalization gaps on the noisy distribution with a noise rate of $0.6$ on CIFAR-100.}
\label{table:gengaps}
\begin{center}
\begin{small}
\begin{sc}
\vskip -0.05in
\begin{tabular}{lcc|c}
\toprule
 
& Models that fit train 100\% & \cellcolor[HTML]{FFFFFF}SAT \cite{huang2020self} & Ours ($\alpha$ = 0.2) \\ 
\midrule
Gen. Gap&\textgreater{}59\% & 12\% & \textbf{6\%} \\ 

\bottomrule
\end{tabular}
\end{sc}
\end{small}
\end{center}

\vskip -0.15in
\end{table*}

Beyond striking empirical generalization performance, there are conceptual reasons to think that self-adaptive frameworks might generalize especially well. By construction, self-adaptive models are able to recognize when examples in the training set are ``confusing'' in the sense that those examples may not be representative of their labeled classes. This is useful when individual labels may be incorrect---but even when there is no label noise, this may also improve generalization by focusing the model on the most representative examples in the training set. Additionally, whereas standard neural networks can achieve 100\% accuracy on the training set, self-adaptive models do not do so by construction, because they change some of the labels in the training set; this implicitly tightens the potential generalization gap, which may give a better pathway to theoretical proofs of generalization.

In the remainder of this section, we present several types of evidence suggesting that self-adaptive models may in fact have strong---potentially provable---generalization performance.

\subsection{Preliminary Experiment: Rademacher Complexity.}

The \textit{Rademacher complexity} $ \mathcal{R}(\mathcal{H})$ of a hypothesis class $\mathcal{H}$ is defined as the expected value of the maximum correlation between a randomly drawn set of labels and any hypothesis in the hypothesis class. It is well-known that a model's generalization error can be bounded by training error plus twice the model's Rademacher complexity with high probability (see, e.g., \cite{shalev2014understanding}). 

\cite{ZhangBHRV17} showed that neural networks can fit arbitrary labels; this implies that the Rademacher complexity of the hypothesis class of neural networks is $1$. By contrast, self-adaptive models cannot fit arbitrary labels---and thus it is plausible that the true Rademacher complexity of the self-adaptive class is low.

\begin{table*}[t!]

\caption{CIFAR-10 with 100\% random labels. Note that test performance is always 10\%.}

\label{table:gengaps100}
\begin{center}
\begin{small}
\begin{sc}
\vskip -0.05in
\begin{tabular}{lcc|c}
\toprule
 
& Models that fit train 100\% & \cellcolor[HTML]{FFFFFF}Mixup \cite{DBLP:conf/iclr/ZhangCDL18} & SAT \cite{huang2020self} \& Ours ($\alpha$ = 0.2) \\ 
\midrule
Train Acc.&100.0\% & 12.1\% & \textbf{10.2\%} \\
Gen. Gap&90.0\% & 2.1\% & \textbf{0.2\%} \\
\bottomrule
\end{tabular}
\end{sc}
\end{small}
\end{center}
\vskip -0.15in
\end{table*}

Indeed, experimental evidence suggests that the empirical Rademacher complexity of self-adaptive training converges to $0$ as the training set increases in size: We mimic the \cite{ZhangBHRV17} experiment by using self-adaptive training to fit random labels, reporting the results in Table~\ref{table:gengaps100}. We run a Wide ResNet 34x10 boosted with self-adaptive training in the same manner as originally done by \cite{huang2020self} on the CIFAR-10 dataset, except with entirely random labels (i.e. almost exactly 10\% of the labels are correct). In each of several runs, the re-weighting process causes the soft labels for every example to converge to almost exactly $(1/c, 1/c, \ldots, 1/c)$, and moreover the predicted class for all test images is identical, and is that of the most common class in the training data. In other words, every example is seen as having a $10\% \pm 0.2\%$ probability of being in each of the 10 classes. This suggests that self-adaptive training's Rademacher complexity is quite low and that arguments that attempt to bound the algorithm's Rademacher complexity might be fruitful---and as we have already noted, such a result would lead to a proof of generalization.

\subsection{Towards Formal Proofs of Generalization.}

Theorizing about the extent to which standard neural network models generalize is especially challenging: since neural networks can both (1) fit arbitrary labels \emph{and} (2) generalize empirically on some datasets, \emph{any proof of generalization must be data-dependent}. This has led to much work proceeding in the direction of showing that specific data manifolds have special properties that support generalization \cite{DBLP:conf/iclr/LyuL20, li2019gradient, brutzkus2018sgd}.

Furthermore, even state-of-the-art neural networks have substantial generalization error. For example, the best test accuracy to-date on CIFAR-100 is 93.6\%---and the highest accuracy of networks trained only on the CIFAR-100 training data is just 89.3\% \cite{cubuk2019autoaugment}, although training accuracy is 100\%.\footnote{Even models trained using billions of images from Instagram as in \cite{mahajan2018exploring} do not get close to 100\% testing accuracy on ImageNet or CIFAR-100.}

If a theorem for neural network generalization were to be found, it would have to take the form
\begin{equation}
\vert \text{\sc training error} - \text{\sc generalization error}\vert \le \Sigma.
\label{eq:Sigma}
\end{equation}
But in the case of CIFAR-100, any theorem pertaining to a currently existing model would need to have $10.7\% \le \Sigma \le 100\%$ in \eqref{eq:Sigma}; it is not clear what sort of theoretical argument would yield an explicit bound on $\Sigma$ that far from $0$.

By contrast, because self-adaptive models have nontrivial error on the training set, the generalization gap for those models can close from both sides. And indeed, the potential value of $\Sigma$ implied by our empirical results is much smaller than the 10.7\% cited above, giving us hope that a formal theorem might be within reach. More broadly, we conjecture that formal results on generalization should be more readily obtainable for classes of models with training accuracy bounded away from 100\%.

\subsection{A Novel Connection to Imbalanced Classes.} 

The \emph{algorithm-dependent VC-dimension}, as defined explicitly in \cite{DBLP:conf/nips/NagarajanK19}, is the VC-dimension of the hypotheses that can be obtained by running the algorithm on some set of labels. A bound on the algorithm-dependent VC-dimension would also imply a generalization bound of $\mathcal{O}(\sqrt{d / m})$, where $d$ is the VC-dimension and $m$ is the size of the training set (\cite{shalev2014understanding}).\footnote{Note that such a bound would not contradict \cite{DBLP:conf/nips/NagarajanK19}, because the training method is not the standard stochastic gradient descent.} Encouraged by the results of the random labels experiment above, we present an argument that the algorithm-dependent VC-dimension of self-adaptive training is actually \emph{extremely limited}. 

Our key idea is inspired by the observation of \cite{DBLP:conf/iclr/SagawaKHL20} that \emph{early stopping fails to fit infrequent classes}. In \cite{DBLP:conf/iclr/SagawaKHL20}, it was noted that empirical risk minimization performed relatively poorly in classifying objects in rare classes, only obtaining worst-class test accuracies of 41.1\% on CelebA. As a supplement, \cite{DBLP:conf/iclr/SagawaKHL20} also tried using early stopping with empirical risk minimization---but  surprisingly, this caused performance to decline even more, bringing worst-case test accuracy down to 25\%.

Our proposed explanation for the surprising phenomenon in \cite{DBLP:conf/iclr/SagawaKHL20} is as follows: It is known that early stopping is resistant to label noise \cite{li2019gradient}. The bound proven in \cite{li2019gradient} is adversarial, in that it shows that early stopping resists any proportion of label noise, even if the noise is chosen adversarially. However, note that \emph{one form of adversarial label noise is the introduction of a rare class}. Intuitively, if a neural network were only given examples corresponding to one class, say, ``airplanes,'' it would learn to classify everything as an airplane. If the network were given exactly one example corresponding to a ``bird'' (alongside all the examples labeled as airplanes), ideally the neural network would learn to classify objects similar to the ``bird'' example as birds. However, noise stability requires that the neural network \textit{should not} learn to classify similar objects as birds, because an ex ante indistinguishable interpretation is that ``all objects are airplanes'' and there was one noisy label. Thus, the objectives of noise resilience and rare-class recognition are conflicting.

As a result, rare-class recognition would seem to be particularly difficult for self-adaptive training. Self-adaptive training repeatedly reinforces model predictions by moving the labels closer and closer to the predictions. If by the time the model begins the label correction phase, most of the predictions for examples in rare classes are still incorrect and fail to match the training labels, those training labels will be shifted towards the dominant class in a ``tyranny of the majority'' effect: almost all of the training labels will become those of the dominant classes.

\begin{figure}[t]
    \centering\vskip -0.15in
    \caption{Class disparities affect self-adaptive training more than they affect vanilla training.\protect\footnotemark}

    \includegraphics[scale = 0.27]{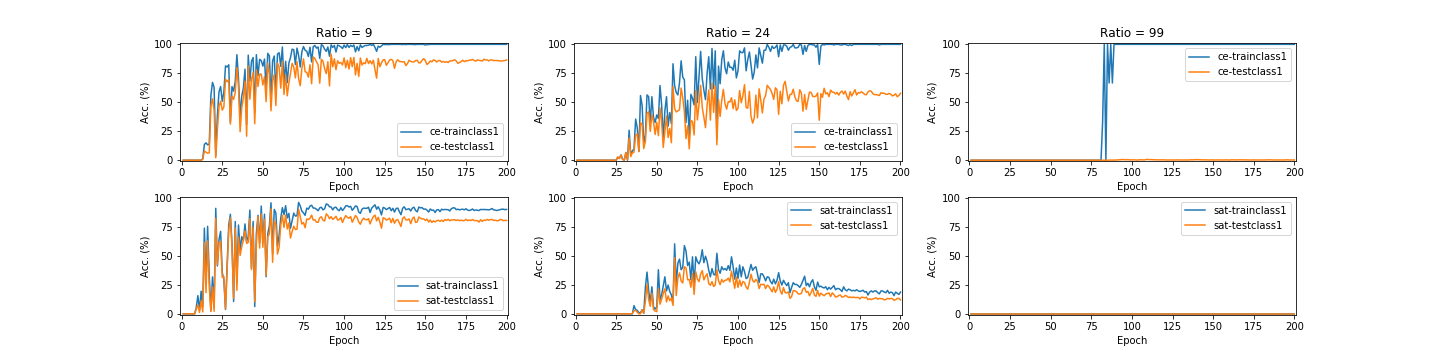}
    \label{fig:imbalanced}
    \vskip -0.25in
\end{figure}

In Figure~\ref{fig:imbalanced}, we show an experiment that runs self-adaptive training on datasets with imbalanced classes. Specifically, the top graphs show the training curves of vanilla cross-entropy training, and the bottom graphs show the training curves of self-adaptive training. For various ratios $r \in \{9, 24, 99\}$, the models are trained on a dataset consisting only of airplanes and automobiles (classes 0 and 1) from CIFAR-10, where the ratio of airplanes to automobiles is $r$ and there are $4500$ airplanes in the training set (the maximum possible). We then report the worst-class accuracy (i.e. accuracy on test set automobiles), as the best-class accuracy is essentially 100\% in all cases. All experiments are done with a Wide ResNet 34x10, following \cite{huang2020self} in all unrelated parameters.

Notice that while self-adaptive training frequently outperforms vanilla training in the presence of balanced classes, these experiments show that it is inferior in the regime of imbalanced classes. In particular, the standard cross-entropy loss gives perfect accuracies for the majority class and decent accuracies for the minority class, while self-adaptive training performs much worse with respect to the minority class. The two variants perform similarly until epoch 60 (when label correction begins), after which the self-adaptive training accuracies begin to decrease (drastically if $r \in \{24, 99\}$) as the model hypothesizes the minority class examples are mislabeled. This is the essence of the tradeoff: cross-entropy training learns the rare class to decent accuracy, while self-adaptive training compromises worst-class accuracy for the sake of robustness to noise.

\footnotetext{Note that, similar to our other results, the generalization gaps on the minority class are much smaller for self-adaptive training than they are for standard cross-entropy training.}

\section{Conclusion}

We have augmented self-adaptive training with mixup, improving calibration---and thus increasing the benefits the model's re-weighting step along lines suggested by theory. The resulting models obtain state-of-the-art accuracies for image classification under label noise. We also noticed strong generalization performance, and provided several threads of reasoning that could lead to formal generalization results for self-adaptive frameworks.

Under conventional wisdom, training a classifier to 100\% accuracy on a training set should improve test performance---which suggests that as far as training goes, longer is better. However, models trained for less time tend to exhibit superior generalization. Self-adaptive models get the best of both worlds; they have better performance than standard training while also taking advantage of shorter training's magnification effects. The key idea, we have argued, is that self-adaptive models can recognize when they are confused and adjust their training progression accordingly. A consequence is that self-adaptive models are likely to be useful beyond settings with label noise, and we expect them to be powerful whenever some examples of classes are more representative than others, which is more or less the generic case.

\begin{ack}
We appreciate the helpful comments of Demi Guo, Daniel Kane, Hikari Sorensen, and members of the Lab for Economic Design (especially Jiafeng Chen,  Duncan Rheingans-Yoo, Suproteem Sarkar, Tynan Seltzer, and Alex Wei).  Kominers gratefully acknowledges the support of National Science Foundation grant SES-1459912, as well as the Ng Fund and the Mathematics in Economics Research Fund of the Harvard Center of Mathematical Sciences and Applications (CMSA). Part of this work was inspired by conversations with Nikhil Naik and Bradley Stadie at the 2016 CMSA Conference on Big Data, which was sponsored by the Alfred~P.\ Sloan Foundation.
\end{ack}

\medskip

\small
\bibliographystyle{plain}
\bibliography{biblio_cln}

\appendix

\newpage \pagestyle{empty}

\section*{Appendix}

\section{The Self-Adaptive Training Algorithm of  \cite{huang2020self}}\label{ap:B}

The algorithm form of self-adaptive training is reproduced below. In particular, label correction appears on lines 6 and 10, and re-weighting appears on lines 8 and 10. In the algorithm, the $\mathbf{t}_i$ represent ``soft labels'' on the examples in the training set, which start out as the (possibly noisy) ``one-hot'' labels. The model trains regularly until epoch $E_s$, when the model begins updating the soft labels based on current predictions. 

\begin{algorithm}[h!]\label{alg:1}
\begin{algorithmic}[1]
\STATE $\{\mathbf{t}_i\}_{n} = \{\mathbf{y}_i\}_{n}$
\FOR{$e = 1$ to $E_s$}

\FOR{$i = 1$ to $m$}
\STATE $\mathbf{p}_i = f(\mathbf{x}_i)$
\IF{$E > E_s$}
\STATE $\mathbf{t}_i = \alpha \times \mathbf{t}_i + (1 - \alpha) \times \mathbf{p}_i$ \COMMENT{\sl LABEL CORRECTION}
\ENDIF
\STATE $w_i = \max_{j} \mathbf{t}_{i, j}$ \COMMENT{\sl RE-WEIGHTING}
\ENDFOR
\STATE $\mathcal{L}(f) = -\frac{1}{\sum w_i} \sum_{i} w_i \sum_{j} \mathbf{t}_{i, j} \log \mathbf{p}_{i, j}$
\STATE Update $f$ by SGD on $\mathcal{L}(f)$.   
\ENDFOR
\end{algorithmic}
\caption{Self-Adaptive Training \cite{huang2020self}}
\label{alg:seq}
\end{algorithm}

\end{document}